\documentclass[conference]{IEEEtran}
\IEEEoverridecommandlockouts
\usepackage{cite}
\usepackage{amsmath,amssymb,amsfonts}
\usepackage{algorithmic}
\usepackage{graphicx}
\usepackage{textcomp}
\usepackage{xcolor}
\usepackage{url}
\def\BibTeX{{\rm B\kern-.05em{\sc i\kern-.025em b}\kern-.08em
    T\kern-.1667em\lower.7ex\hbox{E}\kern-.125emX}}
\begin{document}

    \makeatletter
    \newcommand{\linebreakand}{%
      \end{@IEEEauthorhalign}
      \hfill\mbox{}\par
      \mbox{}\hfill\begin{@IEEEauthorhalign}
    }
    \makeatother

\title{Plagiarism and AI Assistance Misuse in Web Programming: Unfair Benefits and Characteristics}

\author{\IEEEauthorblockN{Oscar Karnalim}
\IEEEauthorblockA{\textit{Faculty of Information Technology} \\
\textit{Maranatha Christian University}\\
Bandung, Indonesia \\
oscar.karnalim@it.maranatha.edu}
\and 
\IEEEauthorblockN{Hapnes Toba}
\IEEEauthorblockA{\textit{Faculty of Information Technology} \\
\textit{Maranatha Christian University}\\
Bandung, Indonesia \\
hapnestoba@it.maranatha.edu}
\and 
\IEEEauthorblockN{Meliana Christianti Johan}
\IEEEauthorblockA{\textit{Faculty of Information Technology} \\
\textit{Maranatha Christian University}\\
Bandung, Indonesia \\
meliana.christianti@it.maranatha.edu}
\linebreakand
\IEEEauthorblockN{Erico Darmawan Handoyo}
\IEEEauthorblockA{\textit{Faculty of Information Technology} \\
\textit{Maranatha Christian University}\\
Bandung, Indonesia \\
erico.dh@it.maranatha.edu}
\and 
\IEEEauthorblockN{Yehezkiel David Setiawan}
\IEEEauthorblockA{\textit{Faculty of Information Technology} \\
\textit{Maranatha Christian University}\\
Bandung, Indonesia \\
2172003@maranatha.ac.id}
\and 
\IEEEauthorblockN{Josephine Alvina Luwia}
\IEEEauthorblockA{\textit{Faculty of Information Technology} \\
\textit{Maranatha Christian University}\\
Bandung, Indonesia \\
2272029@maranatha.ac.id}
}

\maketitle

\begin{abstract}
In programming education, plagiarism and misuse of artificial intelligence (AI) assistance are emerging issues. However, not many relevant studies are focused on web programming. We plan to develop automated tools to help instructors identify both misconducts. To fully understand the issues, we conducted a controlled experiment to observe the unfair benefits and the characteristics. We compared student performance in completing web programming tasks independently, with a submission to plagiarize, and with the help of AI assistance (ChatGPT). 
Our study shows that students who are involved in such misconducts get comparable test marks with less completion time. Plagiarized submissions are similar to the independent ones except in trivial aspects such as color and identifier names. AI-assisted submissions are more complex, making them less readable. Students believe AI assistance could be useful given proper acknowledgment of the use, although they are not convinced with readability and correctness of the solutions.
\\
\end{abstract}

\begin{IEEEkeywords}
plagiarism, artificial intelligence, programming, controlled experiment, computing education
\end{IEEEkeywords}

\section{Introduction}
In programming, code reuse is sometimes promoted for efficiency \cite{Haefliger2008Code}. Nevertheless, such reuse should be acknowledged either as a comment or some text in the documentation. This does not only make the owner(s) of the reused code feel appreciated but also ensure the actual proportion of work contribution.

Plagiarism happens when code is reused without proper acknowledgment \cite{Fraser2014Collaboration}. Instructors typically reduce opportunities to do plagiarism by varying the assessment tasks across semesters \cite{Simon2017Designing} or among students \cite{Bradley2020Creative}. It is also possible to add an extra layer of authentication to prove the originality of the work \cite{Halak2016Plagiarism}. Students who are involved in plagiarism and collusion are identified with the help of automated similarity detectors such as MOSS\footnote{\url{https://theory.stanford.edu/~aiken/moss/}} and SSTRANGE \cite{Karnalim2023Maintaining}.

Instructors need to inform students about plagiarism, including their expectation about the matter in their classes \cite{Simon2018Informing}. This can reduce the incidence of plagiarism and collusion resulted from misrationalization. Sometimes, such information is reminded via automated tools either as general information \cite{Tsang2018Experiential}, simulated similarities \cite{Karnalim2020Disguising}, actual similarities \cite{Le2013Educating}, or a combination of those \cite{Karnalim2022Educating}.

Another way to prevent plagiarism is to reduce student pressure to cheat \cite{Albluwi2019Plagiarism}. Instructors can either incentivize early submissions \cite{Karnalim2023Gamification} or scale down a large assessment to small ones \cite{Allen2018Weekly}. 

Recent advances on artificial intelligence (AI) \cite{chen2022two} complicates identification of plagiarism. Some students might use paraphrasing tools (e.g., Quillbot\footnote{\url{https://quillbot.com/}}) to disguise their misbehavior. They might also use generative AIs such as ChatGPT\footnote{\url{https://openai.com/blog/chatgpt}} to inappropriately help them completing assessments. Paraphrasing tools can be handled by improving the accuracy of existing similarity detectors however identifying AI-assisted submissions can be challenging. Conventional comparison across submissions is not applicable as such an AI-assisted submission is typically unique. 

In programming, the issue exacerbates with the development of many programming languages. Our observation on publicly accessible similarity detectors listed in another study \cite{Blanchard2022Stop} shows that most of the detectors only focus on well-established languages (e.g., C and C++) or the most popular ones (e.g., Java and Python \cite{Simon2018Language}). Web programming languages such as HTML, JavaScript, and CSS are only covered by few similarity detectors, though such languages are quite common in academia and industry.

We are interested to develop a similarity detector and AI-assistance detector for web programming. We also plan to present a guideline about how to integrate both detectors in the educational context. Hence, there is a need to know unfair benefits for students who are involved in plagiarism and AI assistance misuse. It is also important to know the characteristics of such submissions.

In response to the aforementioned gaps, we conducted a controlled experiment at which participants would be asked to complete a web programming task independently, with a submission to plagiarize, and with the help of AI assistance. Their marks and completion time were compared and analyzed. They were also expected to report how they had disguised their act of plagiarism and how they had used AI assistance. Due to recency of AI assistance in completing assessment tasks, we also asked the participants' perspective about the matter.
To the best of our knowledge, this is the first study covering such aspects.

\section{Method}

To assess unfair benefits and characteristics of plagiarism and AI assistance misuse in web programming, a controlled experiment was conducted with tutors (had tutoring experience in the web programming course) and high performing students (their final marks for the course were no less than B+). Although they were unlikely to engage in academic misconducts, they often assessed other students' works and engaged with the students. They knew about how to plagiarize web programming submissions and how to use AI assistance from general student perspective. 

Sixteen participants were randomly divided into two groups (Group 1 and Group 2) and asked to participate in three sessions. Each session took two hours and participants were asked to record their completion time per session. Participation to the experiment was entirely voluntary and each participant would be rewarded by money (around \$5 which is worth for one day meal in Indonesia).

The first session asked students to complete a web programming task independently (without the help of ChatGPT or a submission to plagiarize). For Group 1, the task was taken verbatim from the final test of 2022 web programming offering in our major (referred as FT). Participants were asked to develop a simple yet interactive website to order foods and drinks (minimum five items). The website should have basic HTML, list, table, form, CSS, JavaScript, PHP (to respond form requests), and JSON (to store data). Group 2 completed similar task but in a different domain: tutoring schedules (referred as FT-ALT). 

The second session asked students to complete a web programming task with ChatGPT (a common example of AI assistance). For both groups, the tasks were swapped. Group 1 completed FT-ALT while Group 2 completed FT. Any interactions with ChatGPT were manually recorded and reported.

The third session asked students to complete a web programming task with a submission to plagiarize. Each Group 1 participant was paired with one of Group 2. They were then asked to swap their submissions from the first session. Group 1 participants got submissions of Group 2 participants from the first session and vice-versa. They then plagiarized their assigned submissions with limited changes in program semantic (assuming many students involved in plagiarism were the low performance ones \cite{Joy1999Plagiarism}). Similar to the second session, any changes in the code were recorded and reported.

At the end of the third session, participants were asked about what factors that might be useful to differentiate AI-assisted submissions from the original ones. We also asked about their perspective of AI assistance in completing programming assessments. The survey questions were adapted from another unpublished study and they were:
\begin{itemize}
    \item Q01: I believe AI assistance can help students who are not proficient in programming to complete programming tasks
    \item Q02: I believe AI-generated code can be easily understood by students who are not proficient in programming
    \item Q03: I believe students who are not proficient in programming are likely to use AI assistance to complete programming tasks
    \item Q04: If I was a tutor, I would be able to differentiate AI-generated code from original submissions
    \item Q05: I believe with the correctness of AI-generated code
    \item Q06: I recommend the use of AI assistance for completing programming tasks so long as it is properly cited in the comments and/or the documentation
    \item Q07: I recommend the use of AI assistance for completing general tasks so long as it is properly cited in the comments and/or the documentation
\end{itemize}

All questions should be responded in 5-point Likert scale from strongly disagree (1) to strongly agree (5). Q05-Q07 had additional follow-up questions about reasons for the responses. 

Unfair benefits were measured based on test marks (0-100) and completion time (minutes). Plagiarism and AI assistance misuse got more unfair benefits if involved participants did not get less test marks but could complete the tests faster than those who did the tests independently. Marking was performed by the instructor of 2022 web programming offering.
For plagiarism, submissions from the first session were compared to those from the third session per test (either FT or FT-ALT). The former refers to original submissions while the latter refers to plagiarized submissions. A two-tailed unpaired t-test with 95\% confidence rate were employed to observe statistical significance.
For AI assistance misuse, it is essentially the same except that submissions from the first session (original) were compared to those from the second session (AI-assisted). Further, the t-test was the paired one.

Characteristics of plagiarized submissions and AI-assisted submissions were qualitatively summarized from participants' reports about how they had disguised their act of plagiarism (the third session) and how they had used AI assistance (the second session). Exclusive to AI-assisted submissions, the findings were complemented with participants' responses about what factors that might be useful to differentiate AI-assisted submissions from the original ones.

Participants' perspective about AI assistance was quantitatively analyzed with any free-text responses qualitatively summarized.

\section{Results and Discussion}

\subsection{Unfair Benefits of Plagiarism and AI Assistance}
\label{sec:unfair}

Our experiment shows that for both tests (FT and FT-ALT), plagiarizing participants had comparable test marks to those who completed the tests independently. This also applied to AI-assisted participants. Although Figure \ref{fig:marks} shows that the tests marks were not the same, the differences were not statistically significant.

\begin{figure}[h]
\centering
\includegraphics[width=0.48\textwidth]{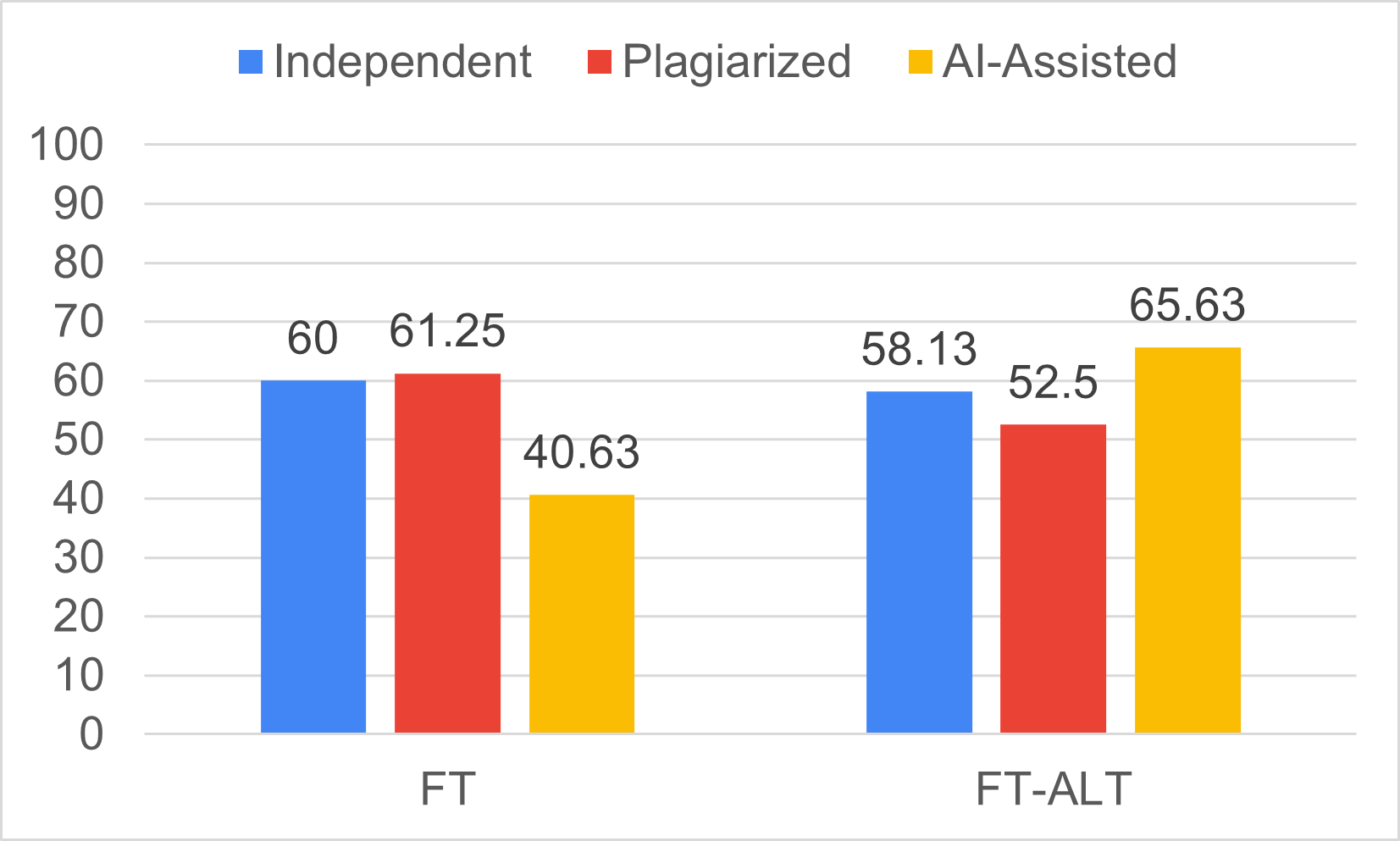}
\caption{Test marks for independent work, plagiarized work, and AI-assisted work}
\label{fig:marks}
\end{figure}

Nevertheless, Figure \ref{fig:time} shows that plagiarizing participants spent less time in completing both tests (only around 20\% of time spent by independent participants). The differences were statistically significant as their p-values $<$ 0.001. AI-assisted participants also spent less completion time though it was still longer than that of plagiarizing participants. They only spent half of the time compared to independent participants and the differences were statistically significant (p-values $<$ 0.01). AI assistance was more difficult to use as the generated code needed to be understood before embedded to one's work. Further, in some cases, the generated code was not completely correct.

\begin{figure}[h]
\centering
\includegraphics[width=0.48\textwidth]{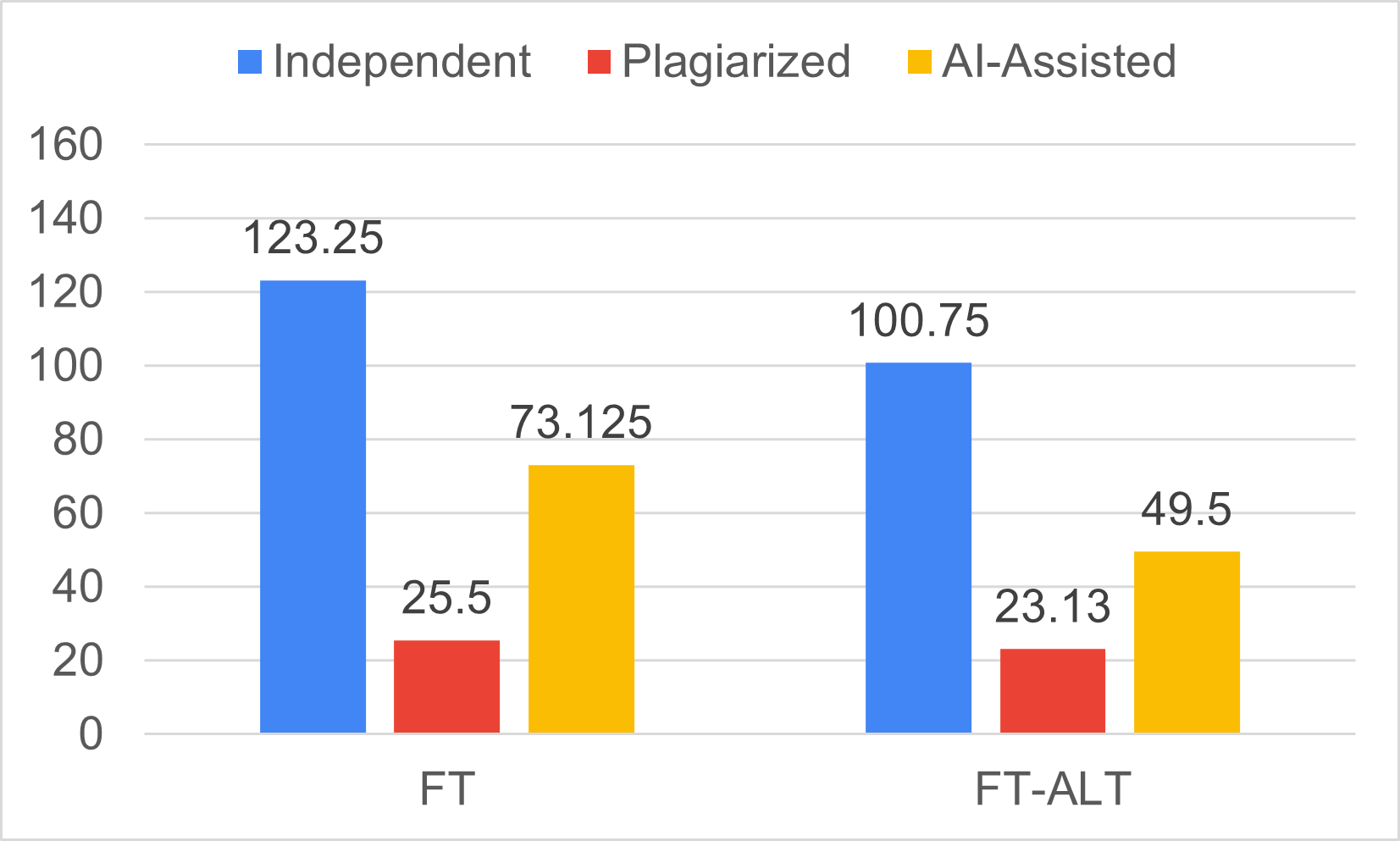}
\caption{Completion time (minutes) for independent work, plagiarized work, and AI-assisted work}
\label{fig:time}
\end{figure}

Plagiarism and AI assistance misuse seemed to have unfair benefits as involved students could get comparable test marks with less effort and shorter completion time. This was quite concerning given that the time differences were substantial and larger than 50\%. There is a need to develop automated tools to detect both misconducts in web programming courses, especially the similarity detector as plagiarizing students spent much less time for completing tests.

\subsection{Characteristics of Plagiarized submissions}

Based on participants' reports, Table \ref{tab:disguises} identified 16 modifications they used to disguise their plagiarized code. The most common disguises were about color change (9) and identifier renaming (9). 
For color change, participants typically chose components that substantially affect the visual layout (e.g., background or body). The new color usually contrasted with the original one (e.g., dark to light or vice-versa).
For identifier renaming, participants changed IDs and class names of HTML elements, including how they were affected by existing JavaScript and CSS files. Some of them updated JavaScript and PHP variables. 

\begin{table}
  \caption{Disguises for Plagiarized Work}
  \centering
  \label{tab:disguises}
  \begin{tabular}{|l|l|c|}
  \hline
  \textbf{ID} & \textbf{Disguise} & \textbf{\# of Participants} \\
  \hline
    D01 & Color change & 9 \\
    D02 & Identifier renaming & 9 \\
    D03 & HTML text change & 7 \\
    D04 & Constant change & 4 \\
    D05 & HTML repositioning & 4 \\
    D06 & Minor HTML restructuring & 3 \\
    D07 & File renaming & 3 \\
    D08 & Component removal & 2 \\
    D09 & Whitespace change & 2 \\
    D10 & File merging and splitting & 1 \\
    D11 & Font style change & 1 \\
    D12 & HTML metadata change & 1 \\
    D13 & Text realignment & 1 \\
    D14 & Comment change & 1 \\
    D15 & Resource change & 1 \\
    D16 & Code reordering & 1 \\
  \hline
 \end{tabular}
\end{table}

HTML text could also be disguised (7) as it was easy to change. Modifications included replacing words with their synonyms, paraphrasing, translating to another human language, and replacing JSON data. Another component that could be disguised was constants (4), like changing default username and password.

HTML repositioning (4) included changing margin, padding, border, and CSS floating properties of HTML elements. Some participants combined that with minor HTML restructuring (3) by either changing the order or the content of HTML elements.

File renaming was performed by three participants. It was quite trivial so long as the linking mechanism across files were maintained. Component removal was performed by two participants focusing on subtle components (e.g., links) and directory structure. Whitespace change was performed by two participants at code level; it did not really affect the visual layout. 

Other disguises were performed by only one participant. Although they were still important for consideration, they could be less prioritized. The disguises were file merging and splitting (e.g., replacing inline CSS to external CSS), font style change, text realignment, comment change, resource change (mostly images), and code reordering (typically happens in JavaScript at function level and in CSS at selector level). 

Plagiarized submissions typically shared the same program semantic with the original ones. However, they might be disguised with some changes in color, identifier names, HTML text, constants, and HTML position. A similarity detector for web programming should be able to nullify such disguises, but to still prioritize verbatim copying to be reported. 

\subsection{Characteristics of AI-Assisted submissions}
\label{sec:AI}

According to participants' reports, they typically started using ChatGPT by breaking down the task to several sub-tasks and asked ChatGPT to generate code per sub-task.
For each generated code, participants embedded that to their own work and updated it if necessary (mostly about resolving conflicting IDs and class names).
In some cases, the code was not relevant or too difficult to integrate. The participants then decided to rewrite it by themselves. This was a reason why many participants did not use ChatGPT for completing the whole tasks.
Some participants simply copied the task instructions and pasted them on ChatGPT.
Unfortunately, it did not provide acceptable results.

To illustrate how our participants had used ChatGPT to complete the tasks, a participant reported the steps as follows:
\begin{enumerate}
    \item Asked ChatGPT to ``create a login HTML page that is featured with username and password check''
    \item Copied the generated code and used that as the basis of work
    \item Asked ChatGPT to ``provide CSS for the HTML page''
    \item Copied the generated code and resolved conflicting class names
    \item Asked about ``how to integrate PHP for checking username and password''
    \item Copied the generated code partially
    \item Asked ChatGPT to ``generate a JSON file for five food items''
    \item Downloaded the JSON file and integrated that in the work
    \item Completed the work independently
\end{enumerate}

The reports indicated that ChatGPT and AI assistants were quite complex to use and only provided partial solutions. Slow-paced students might not be able to modify and integrate the generated code to their own work. They might prefer to plagiarize their colleagues' work if accessible as that was more time-efficient (confirmed in Section \ref{sec:unfair}).

We also asked participants how they could differentiate AI-generated submissions from the original ones. Generally, AI-generated submissions were more complex, employing uncommon syntax and libraries especially for JavaScript and PHP. The submissions might also less readable as some identifier names were generic. AI-assistance detector for web programming should either consider anomaly across submissions (as AI-generated submissions often had uncommon syntax and libraries) or code quality (as AI-generated submissions tended not to have high code quality). 

\subsection{Perspective about AI assistance}

As shown in Figure \ref{fig:perspective}, participants agreed that AI assistance could help slow-paced students in completing programming tasks (Q01 with 3.75 score). Although participants were aware that AI-generated code was not really understandable (Q02 with 2.56 score), they agreed that slow-paced students might still use it for seeking help (Q03 with 3.88 score). 

\begin{figure}[h]
\centering
\includegraphics[width=0.48\textwidth]{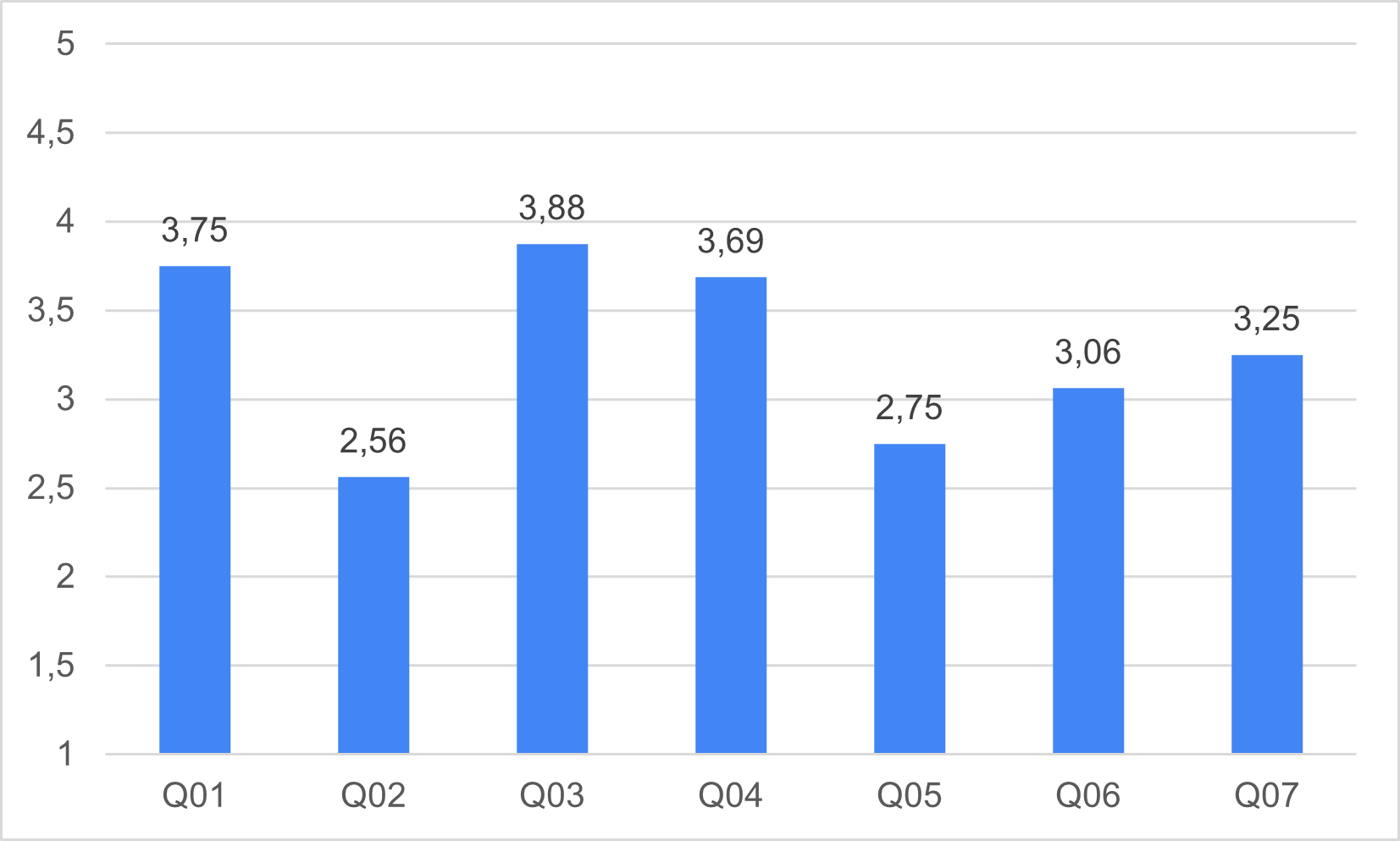}
\caption{Perspective of AI assistance in 5-point Likert scale: 1 refers to strongly disagree, 3 refers to neutral, and 5 refers to strongly agree}
\label{fig:perspective}
\end{figure}

If participants were tutors, they were able to differentiate AI-generated code from the original ones (Q04 with 3.69 score). AI-generated code tended either to be uncommon or not to have high code quality (confirmed in Section \ref{sec:AI}).

Participants did not really believe the correctness of AI-generated code (Q05 with 2.75 score). Based on their follow-up responses, the generated code sometimes did not work or integrating the code needed a lot of work. Both were confirmed in Section \ref{sec:AI}.

Participants were fairly neutral about recommending the use of AI assistance for completing both programming and general tasks given proper acknowledgment (Q06 with 3.05 score and Q07 with 3.25 score). AI assistance might be helpful to deal with simple and common tasks, but it was not really helpful for the complex ones. Students needed to be able to break down a complex task to smaller sub-tasks. In programming, the issue exacerbated with the fact that the generated code was more complex than it should be. Further, it was less readable and might contain errors. All were confirmed in Section \ref{sec:AI} and Q02 Likert score. 

\section{Conclusion}

We presented a controlled experiment to measure unfair benefits and characteristics of plagiarism and AI assistance in web programming tasks. Our findings show that with plagiarism, students might get comparable test marks to those who has completed the test independently with a shorter amount of time. This also applies for those assisted with AI except that the completion time is longer; the generated code needs to be understood and integrated to one's own work.

Plagiarized and independent submissions usually share the same program semantic. Some common disguises are color change, identifier renaming, HTML text change, constant change, and HTML repositioning. AI-assisted submissions tend to have complex syntax and libraries. Further, it might use general identifier names.

Students agreed that AI assistance could be helpful though the generated solutions were not really understandable. They were neutral about the correctness of AI-generated code and recommending the use of AI assistance for completing tasks given proper acknowledgment.

Our study has at least three limitations.
First, the experiment was not performed with real perpetrators of plagiarism and AI assistance misuse given that such perpetrators were unlikely to be interested in helping the authors in developing tools that could detect their misbehaviors.
Second, the number of participants was reasonably small. Participants were limited to tutors and high performing students who often assessed other students works and engaged with them. 
Third, the sequence order of steps in the controlled experiment might affect the result of the study. Reconducting the experiment with different steps might strengthen the findings.

For future work, we plan to develop a similarity detector that can handle common disguises in web programming. We are also interested to develop an AI-assistance detector for the same domain that can rely on code anomaly or code quality.

\section*{Acknowledgment}
To Institution of Research and Community Services, Maranatha Christian University, Indonesia for funding the research.

\bibliographystyle{IEEEtran}
\bibliography{IEEEabrv,ref}

\end{document}